\documentclass[runningheads]{llncs}
\usepackage{cite}
\usepackage[T1]{fontenc}
\usepackage{multirow}
\usepackage{graphicx}
\usepackage{booktabs}
\usepackage{url}

\usepackage{array}

\usepackage{relsize}
\newcommand{\medium}{\fontsize{8.3}{12}\selectfont}

\usepackage{color}

\begin{document}
\title{Evaluation and improvement of Segment Anything Model for interactive histopathology image segmentation}
\titlerunning{SAM for interactive histopathology image segmentation}

\author{SeungKyu Kim, Hyun-Jic Oh, Seonghui Min \and Won-Ki Jeong\thanks{Corresponding author: wkjeong@korea.ac.kr}}

\authorrunning{SeungKyu Kim et al.}
\institute{Korea University, College of Informatics, \\
Department of Computer Science and Engineering \\
\email{wkjeong@korea.ac.kr}
}
\maketitle              
\begin{abstract}
\label{1_abstract}
With the emergence of the Segment Anything Model (SAM) as a foundational model for image segmentation, its application has been extensively studied across various domains, including the medical field. 
However, its potential in the context of histopathology data, specifically in region segmentation, has received relatively limited attention. 
In this paper, we evaluate SAM's performance in zero-shot and fine-tuned scenarios on histopathology data, with a focus on interactive segmentation. Additionally, we compare SAM with other state-of-the-art interactive models to assess its practical potential and evaluate its generalization capability with domain adaptability.
In the experimental results, SAM exhibits a weakness in segmentation performance compared to other models while demonstrating relative strengths in terms of inference time and generalization capability. 
To improve SAM's limited local refinement ability and to enhance prompt stability while preserving its core strengths, we propose a modification of SAM's decoder. 
The experimental results suggest that the proposed modification is effective to make SAM useful for interactive histology image segmentation. 
The code is available at \url{https://github.com/hvcl/SAM_Interactive_Histopathology}
\end{abstract}

\section{Introduction}
\label{2_introduction}
\begin{figure}[t]
    \centering
    \includegraphics[width=0.98\textwidth]{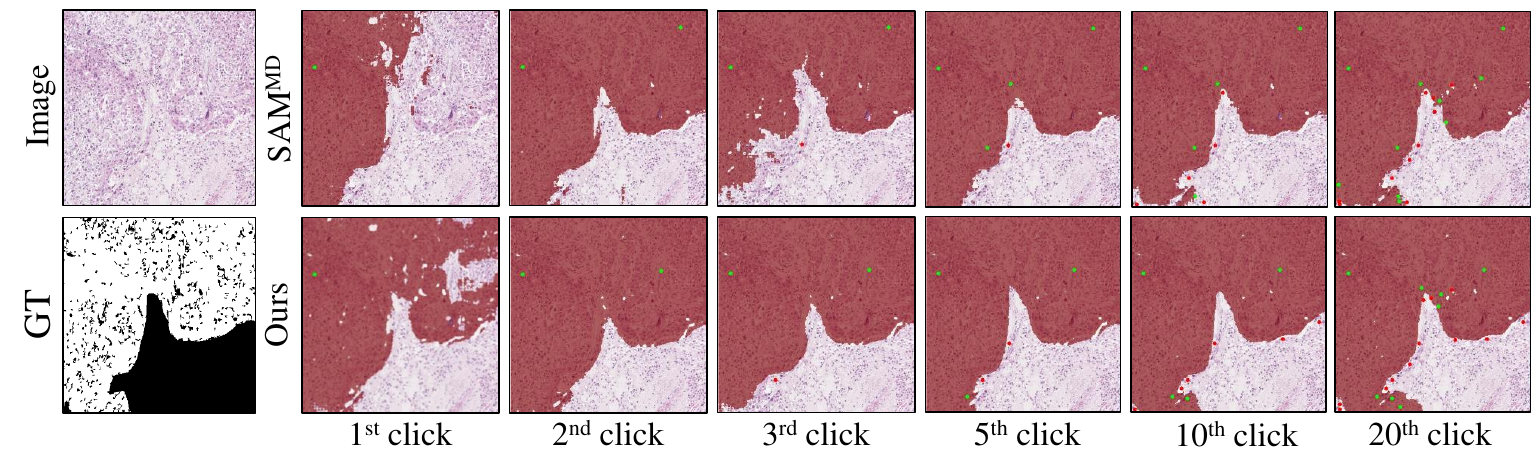}
    \caption{
    Comparison between SAM\textsuperscript{MD} and ours. The green and red points represent positive and negative clicks, respectively. 
    Each click is automatically generated on the error region in the previous prediction.
    }
    \label{fig:SAM_MD_vs_ours}
\end{figure}

Tumor region segmentation in whole slide images (WSIs) is a critical task in digital pathology diagnosis. 
Numerous segmentation methods have been developed in the computer vision community that perform well on objects with clear edges~\cite{graphcut, levelset, randomwalks}. 
However, as shown in Fig.~\ref{fig:SAM_MD_vs_ours}, tumor boundaries in histopathology images are often indistinct and ambiguous. As a result, directly applying conventional image segmentation methods to WSIs tends to yield unsatisfactory results.
In recent years, deep learning advancements have shown promising outcomes in medical image segmentation~\cite{UNet} when sufficient training labels are available. 
Nevertheless, even fully-supervised models have room for improvement due to potential disparities between training and inference imaging conditions, making generalization challenging.

\vspace{10pt}

Recently, the Segment Anything Model (SAM)~\cite{SAM}, a large promptable foundation model that allows user interaction, has gained considerable attention as a general image segmentation model, eliminating the need for task-specific annotation, training, and modeling.
Inspired by this, the primary motivation behind our work stems from the notion that a foundation model like SAM holds promise as a general-purpose segmentation model, which can also be applied to histopathology image segmentation without domain-specific training.
Several early attempts have been made to employ SAM in medical image segmentation~\cite{AutoSAM, MedSAM, DeSAM}. 
However, the exploration of SAM's potential for interactive segmentation has been limited, with only subjective criteria-based prompts being utilized.
Another motivation of our work is that supervised training may not generalize well to various target images during inference. 
Hence, utilizing interactive segmentation to modify the segmentation model's results could prove to be a practical and effective strategy for improving segmentation performance.
While numerous interactive segmentation methods exist in computer vision~\cite{RITM,focalclick,simpleclick}, only a few works have been proposed specifically for interactive histopathology image segmentation~\cite{deepscribble,cgam} so far.

In this study, we investigate the potential of SAM for tumor region segmentation in histopathology images using a click-based interactive approach. 
Our primary objectives are to answer the following questions: 1) Can SAM be directly (zero-shot) applied to interactive histopathology image segmentation tasks? and 2) If not, what is the optimal approach for modifying (or fine-tuning) SAM for interactive histopathology image segmentation?
To address these questions, we conducted extensive experiments on two publicly available datasets: PAIP2019~\cite{paip} and CAMELYON16~\cite{camelyon16}. 
We investigated SAM mainly with point prompts, using a well-established evaluation protocol by Xu et al.~\cite{dios} which mimics human click interaction. 
We conducted a comparison between SAM and other state-of-the-art (SOTA) interactive segmentation methods~\cite{RITM,focalclick,simpleclick}, both with and without dataset-specific fine-tuning. 
We also investigated an efficient strategy to harness the potential of SAM through various fine-tuning scenarios.
The main contributions of our work are several-fold as follows:
\begin{itemize}
    \item We assessed SAM's current capability for zero-shot histopathology image segmentation in the context of interactive segmentation by comparing it against SOTA interactive segmentation algorithms.
    \item We provide insights into the utilization of pretrained weights of SAM by exploring various fine-tuning scenarios. 
    We discovered that SAM requires a prediction refinement strategy for interactive histopathology image segmentation.
    \item We introduce a modified mask decoder for SAM which enhances performance and reduces the fine-tuning cost while retaining the original SAM's high generalization capability and inference speed.
    As a result, we achieved an average reduction of 5.19\% in the number of clicks required to reach the target IoU.
\end{itemize}

\section{Method}
\label{3_method}

\subsection{Overview of Segment Anything Model (SAM)}
Introduced by Meta AI,
SAM~\cite{SAM} is a promptable foundation model for image segmentation trained with the largest segmentation dataset (SA-1B) over one billion masks and 11 million images. 
This model demonstrates significant zero-shot performance in natural image domains.
SAM consists of three main components: 
An Image Encoder (IE) is a Vision Transformer (ViT)~\cite{ViT, mae}-based encoder to extract image features, 
a Prompt Encoder (PE) encodes various types of prompts such as points, bounding boxes, masks, and texts,
and a lightweight Mask Decoder (MD) maps image embedding and prompt embeddings to segmentation results.

\begin{figure}[t]
\includegraphics[width=0.99\linewidth]{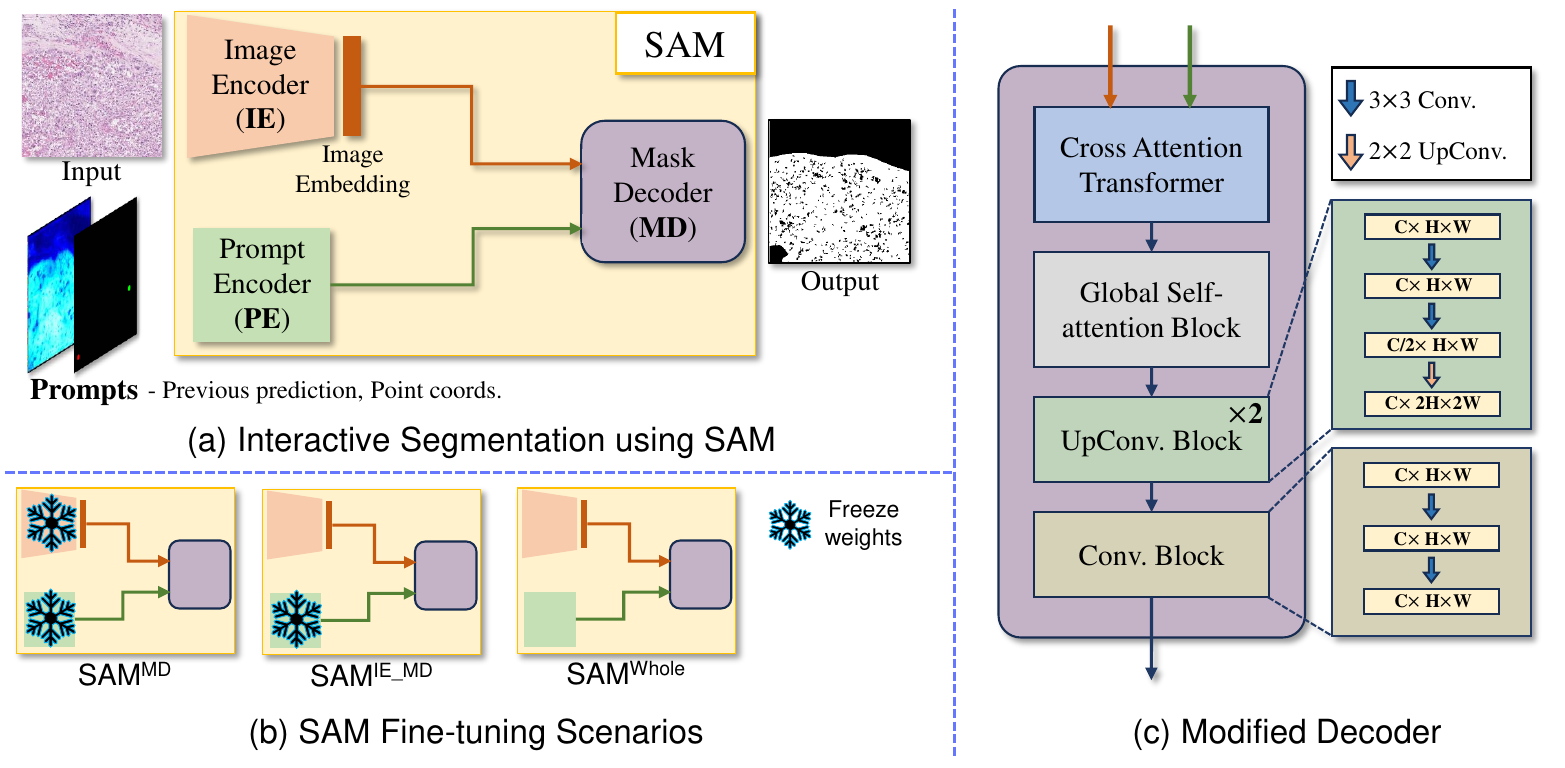}
\caption{
    (a) An illustration of interactive segmentation process using SAM. Image Encoder (IE) extracts features from input patch and Prompt Encoder (PE) encodes prompts which are previous prediction as mask prompt and point coordinates as point prompts.
    (b) We set up three SAM fine-tuning scenarios. 
    SAM\textsuperscript{MD} freezes IE and PE, training only MD.
    SAM\textsuperscript{IE\_MD} freezes PE and trains IE and MD. 
    SAM\textsuperscript{Whole} trains the entire SAM.
    (c) Modified mask decoder architecture to improve local refinement capability. 
    Compared to original MD, we exclude dot-product between the final output token and image embedding. 
    Also, we use global self-attention block, and deeper decoder layers.
    }
\label{fig:SAM_finetuning_modified_decoder}
\end{figure}

\subsection{SAM Fine-tuning Scenarios}
\label{subsec:SAM_components_analysis}
We set up three scenarios for fine-tuning SAM to understand the impact of each component in SAM on the performance and explore an efficient method for utilizing SAM in the interactive segmentation(see Fig.~\ref{fig:SAM_finetuning_modified_decoder} (a) and (b)).
First, we train only the lightweight mask decoder (SAM\textsuperscript{MD}) by freezing the pretrained image encoder and prompt encoder.
Second, we train IE and MD (SAM\textsuperscript{IE\_MD}) to investigate the influence of PE on the model's generalization ability. 
Third, we train the whole model (SAM\textsuperscript{Whole}) for comparison to other scenarios.

To fully utilize the pretrained weights of SAM's ViT-based image encoder, we resized input patches to a size of 1024$\times$1024 and restored the output predictions to match the original input patch size.
In the training process, we employed the click guidance scheme by Sofiiuk et al.~\cite{RITM} for automatic point prompts generation.
The click guidance scheme uses random sampling for the first iteration and samples subsequent clicks from the error regions of the previous predictions, thereby better resembling real-world user interaction. 
Moreover, we use the previous prediction as a mask prompt to improve the model performance as shown in~\cite{iterative_train}. 
We employed the Normalized Focal Loss~\cite{RITM}, known for faster convergence and better accuracy compared to Binary Cross-Entropy (BCE) as explained in~\cite{simpleclick, focalclick, pseudoclick}.

\subsection{Decoder Architecture Modification}

\begin{figure}[t]
\includegraphics[width=0.99\linewidth]{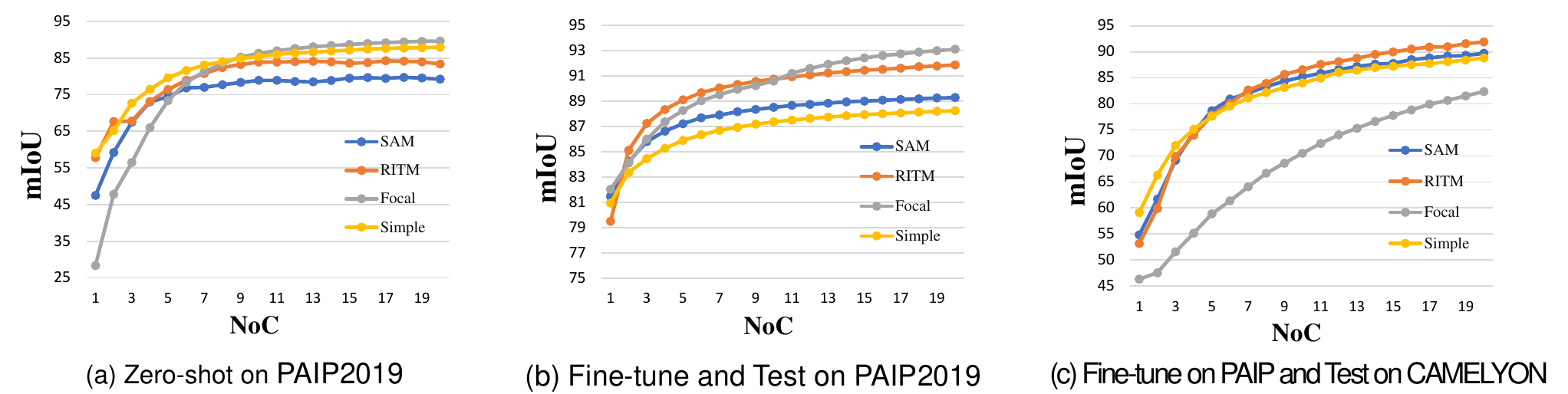}
\caption{
    (a) Zero-shot mIoU scores per click of each model on the PAIP2019 dataset.
    (b) mIoUs per click after fine-tuning on the PAIP dataset. 
    (c) mIoUs per click on the CAMELYON16 $\times$10 dataset for models trained on the PAIP dataset.
    }
\label{fig:fig3_plot}
\end{figure}

The graph in Fig.~\ref{fig:fig3_plot} (a), and (b) show the zero-shot and fine-tuned performance with mean Intersection over Union (mIoU) per interaction of SAM and SOTA interactive models on PAIP2019~\cite{paip} dataset.
SAM demonstrates comparable performance to other models in early iterations, but it struggles in later iterations.
As shown in Fig.~\ref{fig:qual_zero_fine}, SAM without fine-tuning shows weakness in refining predictions locally (i.e., a local modification affects a large area). 
Considering the architecture of the ViT-based SAM image encoder, training the entire SAM requires a longer time and a higher computational cost. 
Also, as shown in Table~\ref{table:zeroshot} and described in Sec.~\ref{subsection:SAM_Finetuning}, we empirically found that including IE in fine-tuning is not always beneficial.

To address this issue, we modified the lightweight decoder in the original SAM as depicted in Fig.~\ref{fig:SAM_finetuning_modified_decoder} (c) to improve local refinement capability and assign prompt stability. 
We add a global self-attention layer to the image embedding after the cross-attention transformer block to enhance the ability to capture the global context across the entire patch. 
Moreover, we deepen the decoder layers during the upsampling process to increase the representational capacity of the decoder.

The upsampling process is described as follows:
We constructed a \textit{UpConvBlock} and \textit{ConvBlock} module. \textit{UpConvBlock} consists of two 3$\times$3 convolution layers and a 2$\times$2 up-convolution layer and \textit{ConvBlock} consists of two 3$\times$3 convolution layers. In the second 3$\times$3 convolution layer of both \textit{UpConvBlock} and \textit{ConvBlock}, a channel reduction is performed, reducing the input channel size by half.
After global self-attention, the image embedding proceeds through two \textit{UpConvBlock}s and \textit{ConvBlock} sequentially. 
We integrated instance normalization layers and \textit{GELU} activation functions between each layer.
Furthermore, to mitigate information loss, we employed shortcut connections within each Block. 
After passing through 3 Blocks, features are transformed into a segmentation map through the 1$\times$1 convolution layer.

\section{Experiments}
\label{4_experiments}

\subsection{Data Description}
We utilized two publicly available WSI datasets, namely PAIP2019~\cite{paip} and CAMELYON16~\cite{camelyon16}. 
The PAIP2019 WSIs were scaled at a 5× magnification, while for CAMELYON16, we used WSIs scaled at both 10× and 5× magnification. 
Subsequently, the WSIs were cropped into patches of size 400×400 pixels. 
These patches were then filtered based on the proportion of tumor area, ranging from 20\% to 80\%, as outlined in~\cite{cgam}. 
Consequently, for the PAIP2019 dataset, we employed 5190, 576, and 654 patches for training, validation, and testing, respectively. 
As for CAMELYON16, we utilized 397 patches at 10× magnification and 318 patches at 5× magnification scales for testing.

\subsection{Implementation Details}
We compared SAM to three SOTA interactive segmentation methods: RITM~\cite{RITM}, FocalClick~\cite{focalclick}, and SimpleClick~\cite{simpleclick}. 
We fine-tuned the pretrained models in PAIP2019 and validated their generalization capabilities in CAMELYON16. 
For a fair comparison, we used the largest pretrained model and the best-performing pretrained weights for each model.
For fine-tuning the parameters, we adhered to the default setting of~\cite{RITM}, wherein we set the initial learning rate to 5e-4. 
Additionally, we decreased the learning rate by a factor of 10 at the 20th and 25th epochs out of the total 30 epochs. 

To assess the zero-shot transfer ability of each model, we conducted inference without training using WSIs. 
Additionally, to evaluate the generalization ability and domain adaptability, we performed inference after fine-tuning using WSIs.
For a fair comparison, we employed an automatic evaluation method used in the previous work
~\cite{LatentDiversity, conditionaldiffusion, BRS, f-brs, dios, 1clickattention}. 
Inference per patch continues until it reaches the target IoU (Intersection over Union).
We set the maximum number of clicks per patch to 20.
As for the evaluation metrics, we used Number of Clicks (NoC), Seconds per Click (SPC), and Number of Fails (NoF). 
NoC represents the average number of clicks required to reach the target IoU, while SPC measures the average time it takes for the model to perform an inference per click. 
NoF represents the number of images that failed to reach the target IoU despite the maximum number of clicks. 
For a more intuitive representation, we divide NoF by the number of testset \textit{n}.
All of evaluations were performed on a single NVDIA RTX A6000 GPU.

\section{Results}
\label{5_Results}
\begin{figure}[t]
\includegraphics[width=0.99\linewidth]{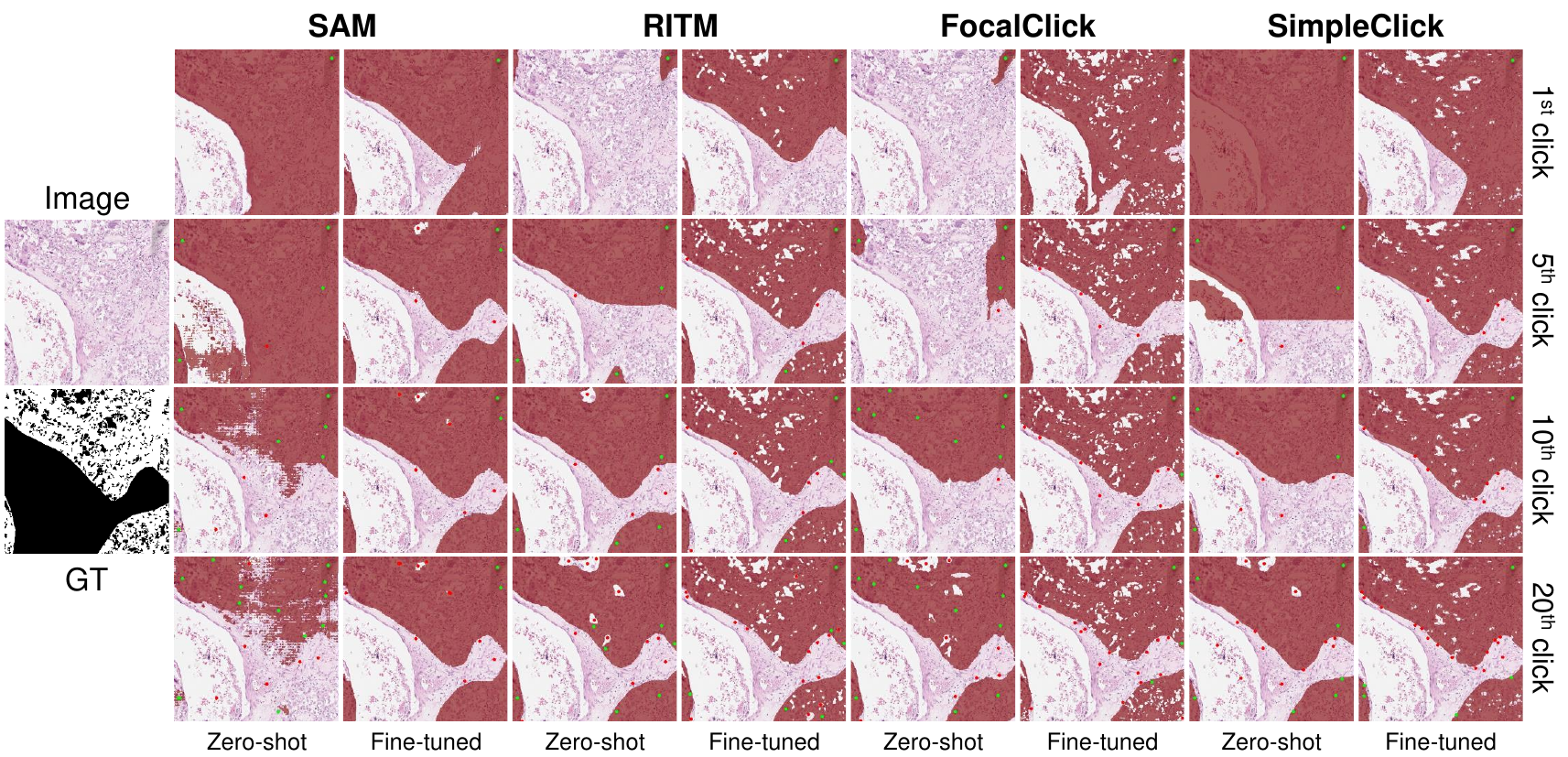}
\caption{Qualitative comparison between SAM\textsuperscript{Whole} and SOTA methods. 
The results show the performance of each method on the PAIP2019 dataset, both in terms of zero-shot and after fine-tuning.}
\label{fig:qual_zero_fine}
\end{figure}
\begin{table}[t]
\centering
\caption{Quantitative Result of Zero-Shot Performance}
\renewcommand{\arraystretch}{1.1}
\setlength{\tabcolsep}{5.3pt}
\label{table:zeroshot}
\begin{tabular}{c|c|ccc|c|cc}
\toprule
\specialrule{0.1pt}{0pt}{0pt}
\multirow{2}{*}{\textbf{Dataset}} & \multirow{2}{*}{\textbf{Method}} & \multicolumn{3}{c|}{NoC@} & \multirow{2}{*}{SPC(s)} & \multicolumn{2}{c}{NoF$/$\textit{n} $@$} \\
\cline{3-5}
\cline{7-8}
\multicolumn{1}{c|}{} & \multicolumn{1}{c|}{} & 80 & 85 & 90 & \multicolumn{1}{c|}{} & 85 & 90 \\
\hline\hline
\multirow{4}{*}{\textbf{\medium{\shortstack{PAIP2019\\ \\($\times$5)}}}}

& \medium{RITM} &7.43& 10.24& 13.00& 0.075 & 0.30& 0.47\\

& \medium{FocalClick}  & 7.37& \textbf{9.89}& \textbf{12.42}& 0.073& \textbf{0.28} & \textbf{0.43} \\

& \medium{SimpleClick} & \textbf{7.21}& 10.16& 13.44& 0.189  & 0.33&0.53\\ 

& \medium{SAM} & 9.13 & 12.10 & 14.73 &  \textbf{0.052} & 0.50 & 0.66 \\ 
\hline

\multirow{4}{*}{\textbf{\medium{\shortstack{CAMELYON16\\ \\ ($\times$5)}}}}

& \medium{RITM} & 6.25 & 7.88& 9.98& 0.078 &0.14&0.24\\

& \medium{FocalClick} & \textbf{3.64}& \textbf{4.83}& \textbf{7.20}& 0.085 &\textbf{0.04}&\textbf{0.14} \\

& \medium{SimpleClick} & 5.26& 6.94& 9.23& 0.187 &0.15&0.25\\ 

& SAM & 5.31& 7.45& 10.38& \textbf{0.053}&0.19&0.36 \\ 
\hline

\multirow{4}{*}{\textbf{\medium{\shortstack{CAMELYON16\\ \\($\times$10)}}}}

& \medium{RITM} & 6.91& 8.19& 10.07& 0.077&0.15&0.22\\

& \medium{FocalClick} & \textbf{4.73}& \textbf{5.89}& \textbf{7.87}& 0.076&\textbf{0.06}&\textbf{0.14} \\

& \medium{SimpleClick} & 5.20& 6.42& 8.55&  0.187&0.11&0.23\\ 

& \medium{SAM} & 6.64& 8.49& 11.03& \textbf{0.053}&0.24&0.37\\ 
\toprule
\specialrule{0.1pt}{0pt}{0pt}
\end{tabular}

\end{table}

\begin{table}[htb!]
\centering
\caption{Quantitative Result of After Fine-tuning on PAIP2019}
\label{table:finetune}
\renewcommand{\arraystretch}{1.05}
\setlength{\tabcolsep}{1.0pt}
\begin{tabular}{c|c|ccc|c|cc}
\toprule
\specialrule{0.1pt}{0pt}{0pt}
\multirow{2}{*}{\textbf{Dataset}} & \multirow{2}{*}{\textbf{Method}} & \multicolumn{3}{c|}{NoC@} & \multirow{2}{*}{SPC(s)} & \multicolumn{2}{c}{{NoF$/$\textit{n} $@$}} \\
\cline{3-5}
\cline{7-8}
\multicolumn{1}{c|}{} & \multicolumn{1}{c|}{} & 80 & 85 & 90 & \multicolumn{1}{c|}{} & 85 & 90 \\
\hline\hline

\multirow{7}{*}{\textbf{\medium{\shortstack{PAIP2019\\ \\ ($\times$5)}}}}

& \medium{RITM} & \underline{2.78}& 4.93& 9.40& 0.075 &0.12&0.33\\

& \medium{FocalClick} & \textbf{2.58}& \textbf{4.54}& \textbf{8.98}&0.071 &\textbf{0.06}&\textbf{0.25}\\ 

& \medium{SimpleClick} & 5.11& 8.33& 12.20&  0.189 &0.32&0.52\\ 

\cline{2-8}
& \medium{SAM\textsuperscript{MD}}& 5.80& 8.93& 12.09&\textbf{0.050} &0.31&0.48\\ 

& \medium{SAM\textsuperscript{IE\_MD}}& 4.53& 7.58& 10.86& \textbf{0.050}&0.29&0.45\\ 

& \medium{SAM\textsuperscript{Whole}}& 4.53& 7.50& 10.95& \textbf{0.052}&0.28&0.46\\ 

\cline{2-8}
& Ours& 4.75& 7.78& 10.85& 0.067 &0.26&0.42\\

\hline
\hline
\multirow{7}{*}{\textbf{\medium{\shortstack{CAMELYON16\\ \\($\times$5)}}}} 
& \medium{RITM} & 3.74\tiny{(-2.51)}& 5.04\tiny{(-2.84)}& 7.50\tiny{(-2.48)}&0.081&\textbf{0.08}&\textbf{0.17}\\

& \medium{FocalClick} & 4.15\tiny{(+0.51)}& 5.57\tiny{(+0.74)}& 8.19\tiny{(+0.99)}& 0.071&\underline{0.10}&0.22 \\

& \medium{SimpleClick} & 3.82\tiny{(-1.44)}& 5.41\tiny{(-1.53)}& 7.94\tiny{(-1.29)}& 0.187&0.15&0.25 \\ 

\cline{2-8}
& \medium{SAM\textsuperscript{MD}}& 4.09\tiny{(-1.22)}& 5.84\tiny{(-1.61)}& 8.31\tiny{(-2.07)}&\textbf{0.050} &0.13&0.24\\ 

& \medium{SAM\textsuperscript{IE\_MD}}& 4.30\tiny{(-1.01)}& 5.64\tiny{(-1.81)}& 7.85\tiny{(-2.53)}&\textbf{0.052}&0.15&0.24\\ 

& \medium{SAM\textsuperscript{Whole}}& \textbf{3.28}\tiny{(-2.03)}& \textbf{4.80}\tiny{(-2.65)}& \textbf{7.06}\tiny{(-3.32)}& \textbf{0.053}&\underline{0.10}&\underline{0.20}\\ 
\cline{2-8}
& Ours& 4.27\tiny{(-1.04)}& 5.59\tiny{(-1.86)}& 8.04\tiny{(-2.34)}&\underline{0.066}&\underline{0.11}&\underline{0.20}\\

\hline
\hline
\multirow{7}{*}{\textbf{\medium{\shortstack{CAMELYON16\\ \\($\times$10)}}}}

& \medium{RITM} & 6.28\tiny{(-0.63)}& 7.65\tiny{(-0.54)}& 9.67\tiny{(-0.31)}& 0.076 &\underline{0.13}&\underline{0.23}\\

& \medium{FocalClick} & 11.82\tiny{(+7.09)}& 13.01\tiny{(+7.12)}& 14.53\tiny{(+7.33)}& 0.076&0.43&0.54 \\ 

& \medium{SimpleClick} & 6.07\tiny{(+0.87)}& 7.44\tiny{(+1.02)}& 9.94\tiny{(+1.39)}&  0.187&0.18&0.28\\ 

\cline{2-8}
& \medium{SAM\textsuperscript{MD}}& \underline{4.88}\tiny{(-1.76)}& 6.68\tiny{(-1.81)}& 9.07\tiny{(-1.96)}& \textbf{0.053}&\underline{0.13}&0.25\\ 

& \medium{SAM\textsuperscript{IE\_MD}}& 7.63\tiny{(+0.99)}& 9.19\tiny{(+0.7)}& 11.81\tiny{(+0.78)}&\textbf{0.049}&0.23&0.38\\ 

& \medium{SAM\textsuperscript{Whole}}& 6.60\tiny{(-0.04)}& 8.10\tiny{(-0.39)}& 10.66\tiny{(-0.37)}& \textbf{0.053}&0.20&0.31\\ 

\cline{2-8}
& Ours& \textbf{4.59}\tiny{(-2.05)}& \textbf{5.92}\tiny{(-2.57)}& \textbf{8.40}\tiny{(-2.63)}& \underline{0.064}&\textbf{0.11}&\textbf{0.20} \\

\toprule
\specialrule{0.1pt}{0pt}{0pt}
\end{tabular}

\end{table}
\subsection{Zero-shot Performance}
\label{subsection:zero_shot}
Table~\ref{table:zeroshot} presents the zero-shot performance of SAM and state-of-the-art (SOTA) interactive segmentation models. 
SAM demonstrates lower performance compared to SOTA models across all three different domains. 
The low NoC and high NoF in the zero-shot setting indicate that SAM lacks the ability to effectively modify the mask to accurately reflect user intent, particularly when faced with challenging images. 
As depicted in Fig.~\ref{fig:qual_zero_fine}, SAM in the zero-shot setting fails to refine local information. 
However, SAM exhibits minimal SPC compared to other models due to its feature extraction being performed only once on the input image, resulting in a significantly faster processing time, over three times faster than SimpleClick.

\subsection{Fine-tuned SAM Performance}
\label{subsection:SAM_Finetuning}
As introduced in Sec~\ref{subsec:SAM_components_analysis}, we compare the three scenarios to verify the impact of each SAM component on fine-tuned performance.
We fine-tuned the entire SAM (SAM\textsuperscript{Whole}), the IE and MD (SAM\textsuperscript{IE\_MD}), and only the MD (SAM\textsuperscript{MD}) using the PAIP2019 dataset, respectively. 
Then, we verify the generalization ability of models at two scales of the CAMELYON16 dataset.
As shown in Table~\ref{table:finetune}, SAM\textsuperscript{MD} yield lower performance on NoC and NoF compared to IE-trained scenarios on the PAIP2019 dataset.
On CAMELYON16, SAM\textsuperscript{MD} exhibited comparable results as IE-trained cases at scale $\times$5 and better performance at scale $\times$10. 
This result demonstrates that the pretrained IE of SAM shows the robustness of feature extraction ability on different data distributions.

\subsection{Comparison between SAM and SOTA Interactive Methods}

As shown in Table~\ref{table:finetune}, SAM still exhibits a notable performance gap compared to RITM and FocalClick on the PAIP2019 dataset. 
SAM requires at least two more NoC compared to RITM and FocalClick. 
SimpleClick exhibited poorer performance compared to SAM\textsuperscript{Whole} and SAM\textsuperscript{IE\_MD}, 
but similar or slightly better performance compared to SAM\textsuperscript{MD}. 
On the CAMELYON16 $\times$5 dataset, SAM\textsuperscript{Whole} exhibited the best performance. 
The numbers in parentheses in Table~\ref{table:finetune} represent the changes in the NoC metric compared to the zero-shot performance in Table~\ref{table:zeroshot}. 
All models showed a decrease in NoC but an increase in FocalClick, indicating that CAMELYON16 $\times$5 has similar data distribution with the PAIP2019 dataset. 
As for the CAMELYON16 $\times$10 dataset, SAM\textsuperscript{MD} performed second best after SAM with a modified decoder (ours).
Furthermore, despite significant differences in data distributions between source and target domain, SAM\textsuperscript{MD} and SAM\textsuperscript{Whole} demonstrated improved performance along with RITM, indicating a higher generalized capability compared to FocalClick and SimpleClick. 
Note that FocalClick and SimpleClick performed poorly compared to the zero-shot approach with FocalClick's significant performance drop
as shown in Fig.~\ref{fig:fig3_plot} (c).
These methods utilize local segmentation which refines the prediction locally.
As the data distribution changes, more iterations of local segmentation are required, 
resulting in performance degradation in both NoC and NoF metrics.

\subsection{Modified SAM Decoder Performance}

As shown in Table~\ref{table:finetune}, our approach exhibits improved performance compared to SAM\textsuperscript{MD} in every dataset.
Moreover, it shows comparable performance in PAIP2019 compared to SAM\textsuperscript{Whole}.
Especially on the CAMELYON16 $\times$10 dataset, our approach demonstrates the best performance among all interactive SOTA models. 
It shows the highest performance in all of NoC and NoF, highlighting its remarkable generalization capability.
However, in PAIP2019, it still shows lower performance compared to SOTA interactive models.
Intuitively, Table 2 may not be enough to say that our approach has a clear advantage over other methods. However, when taking into account a comprehensive range of factors including performance, inferencing speed, training cost, and generalization ability, we argue that our approach is sufficiently compelling.

\section{Conclusion}
\label{6_conclusion}

In this study, we demonstrate the potential of SAM for interactive pathology segmentation.
SAM showed higher generalization capability and notably excelled in terms of inference speed per interaction compared to SOTA interactive models. 
Additionally, a modified decoder with pretrained encoders of SAM can achieve performance comparable to that of the entire SAM fine-tuning. 
In this approach, SAM could efficiently capture user intent and precisely refine predictions, which is crucial in interactive segmentation.
However, SAM exhibits lower performance compared to state-of-the-art interactive models, especially in achieving high IoU scores. 
We plan to develope SAM as a foundational interactive model that works well for all histopathology data (different organs, tissues, etc.).

\subsection*{\textbf{Acknowledgements}}This work was partially supported by the National Research Foundation of Korea (NRF-2019M3E5D2A01063819, NRF-2021R1A6A1A
13044830), the Institute for Information \& Communications Technology Planning
\& Evaluation (IITP-2023-2020-0-01819), the Korea Health Industry Development
Institute (HI18C0316), the Korea Institute of Science and Technology (KIST)
Institutional Program (2E32210 and 2E32211), and a Korea University Grant.

\bibliographystyle{splncs04}
\bibliography{7_refers}

\clearpage
\label{8_supplementation}
\setcounter{table}{0}
\renewcommand{\thetable}{S\arabic{table}}
\setcounter{figure}{0}
\renewcommand{\thefigure}{S\arabic{figure}}

\begin{figure}[htb]
\includegraphics[width=0.99\linewidth]{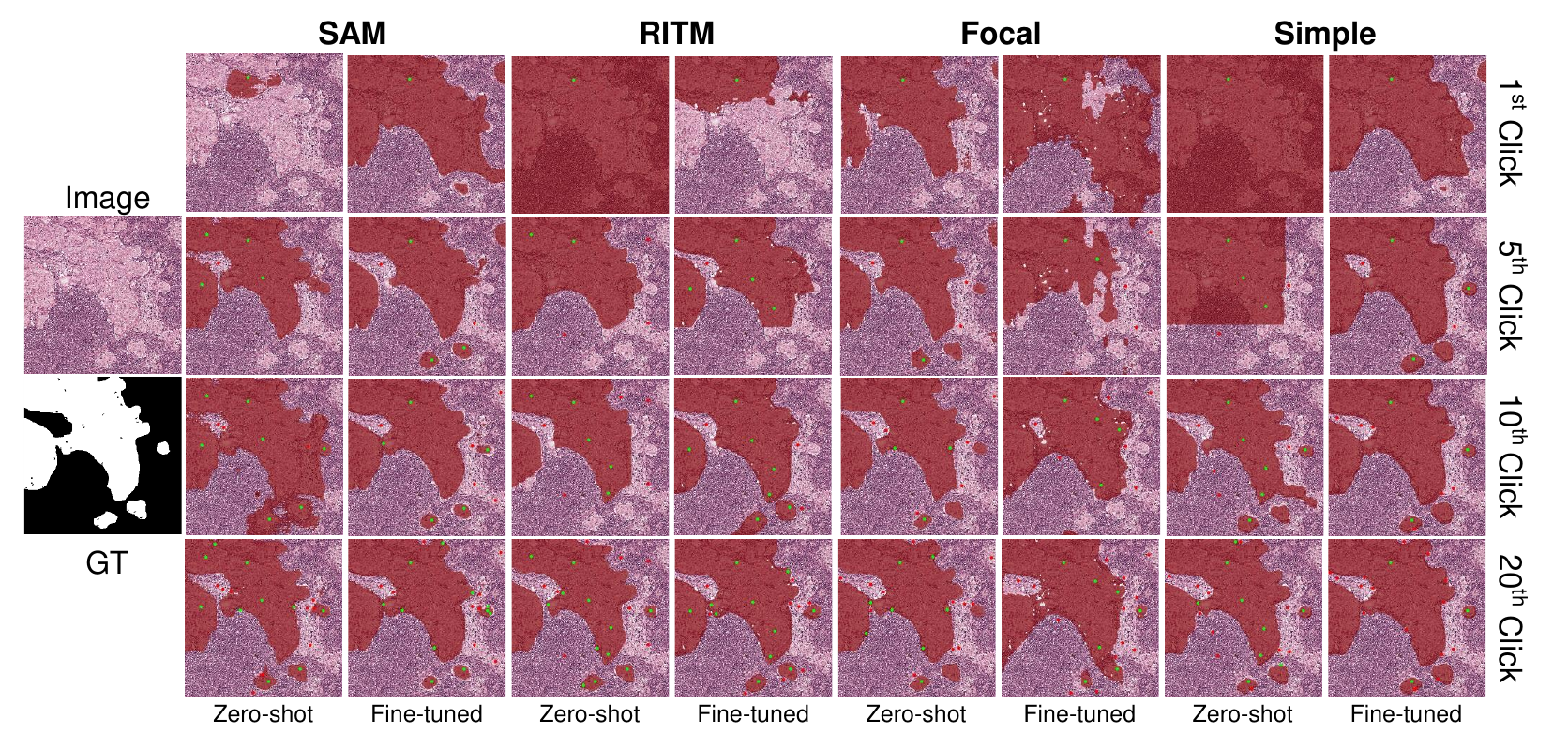}
\caption{Qualitative comparison between SAM\textsuperscript{Whole} and SOTA methods on CAMELYON16 $\times$5 dataset.
The results on each method show the performance in zero-shot and fine-tuned on PAIP2019, respectively.
The segmentation results in each row correspond to different numbers of clicks: one click, five clicks, ten clicks, and twenty clicks. 
Except for FocalClick, all other models demonstrated improved results over the zero-shot, with SAM showing the largest improvement.
}
\label{fig:camX5_qual}
\end{figure}

\begin{figure}[htb!]
\includegraphics[width=0.99\linewidth]{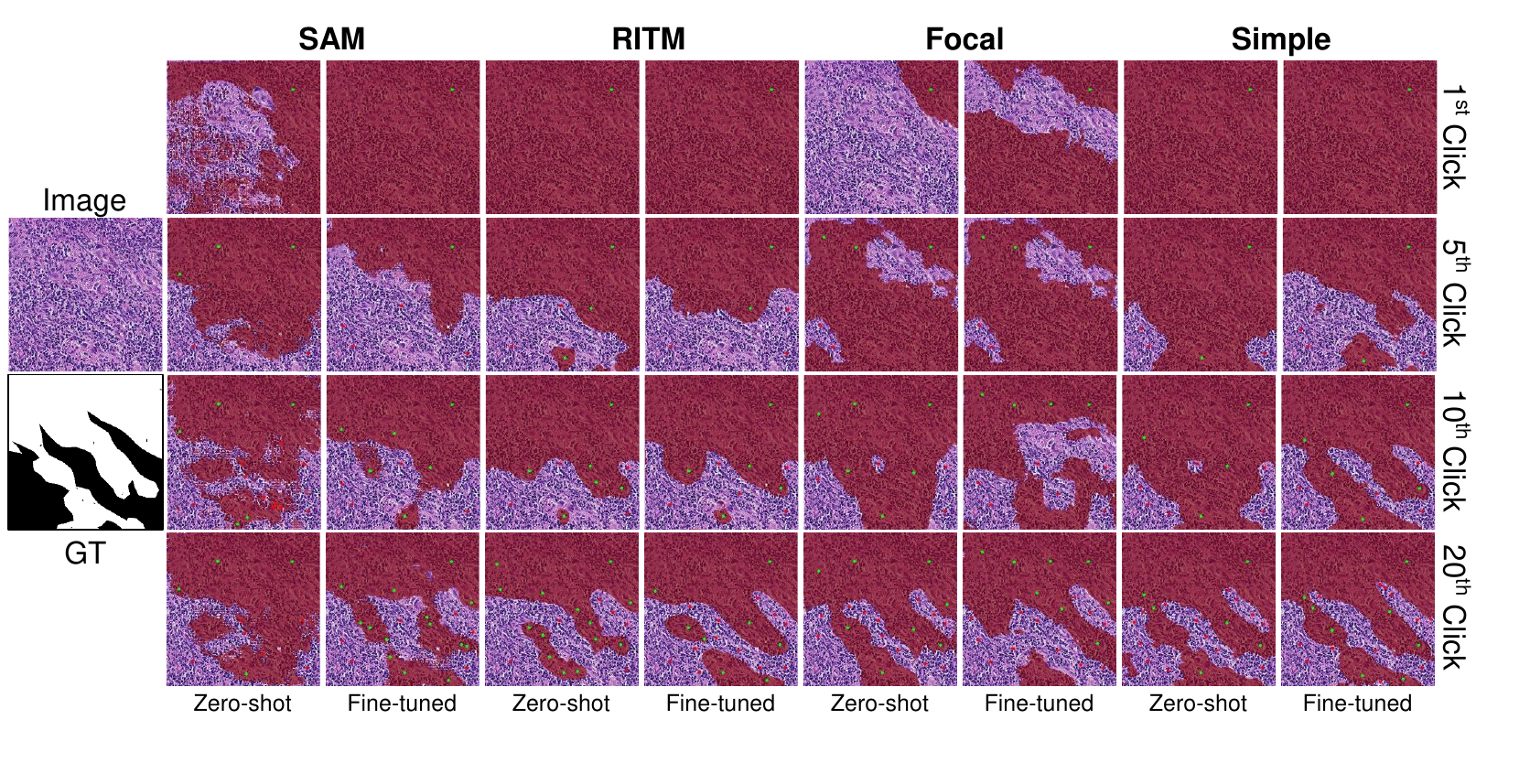}
\caption{Qualitative comparison between SAM\textsuperscript{Whole} and SOTA methods on CAMELYON16 $\times$10 dataset.
The results on each method show the performance in zero-shot and fine-tuned on PAIP2019, respectively.
}
\label{fig:camX10_qual}
\end{figure}

\end{document}